\documentclass[11pt,a4paper]{article}
\usepackage[hyperref]{acl2021}
\usepackage{times}
\usepackage{latexsym}

\usepackage{microtype}

\aclfinalcopy 

\usepackage{epigraph} 
\usepackage{csquotes} 
\usepackage{enumitem} 
\usepackage{multirow}
\usepackage{booktabs}
\usepackage{graphicx}
\usepackage{amsmath}

\newenvironment{myquote}[1]%
  {\list{}{\leftmargin=#1\rightmargin=#1}\item[]}%
  {\endlist}

\newcommand{\af}[1]{\textcolor{blue}{#1}}

\title{What is Multimodality?
}

\author{Letitia Parcalabescu \\ Computational Linguistics \\ Department \\ Heidelberg University \\ 
        \And
        Nils Trost \\ Center for Molecular \\ Biology (ZMBH) \\ Heidelberg University 
        \\ 
        \And
        Anette Frank \\ Computational Linguistics \\ Department \\ Heidelberg University 
        \AND \vspace{-10mm}\\ 
       \small{ \texttt{\{parcalabescu,frank\}@cl.uni-heidelberg.de ~~~ \texttt{trost@zmbh.uni-heidelberg.de}}}
        }
\begin{document}
\maketitle

\begin{abstract}
The last years have shown rapid developments in the field of multimodal machine learning, combining e.g., vision, text or speech. In this position paper we explain how the field uses outdated definitions of \emph{multimodality} that prove unfit for the \emph{machine learning} era.
We propose a new \emph{task-relative} definition of \emph{(multi)modality} in the context of multimodal machine learning that focuses on representations and information that are \emph{relevant} for a given machine learning task. With our new definition of multimodality we aim to provide a missing foundation for multimodal research, an important component of language grounding and a crucial milestone towards NLU.
\end{abstract}

\section{Introduction}

The holy grail of NLP is natural language understanding (NLU). As previously argued, NLU cannot be achieved by learning from text 
alone
\cite{bender-koller-2020-climbing}. Instead, an important step towards NLU is grounding language, especially in sight and sounds \cite{bisk-etal-2020-experience}. 
There is thus great interest in the field of NLP to go beyond the textual modality and to conduct multimodal machine learning (ML) research.

Multimodal ML has made great progress during the last years. Neural architectures are employed in tasks that go beyond single modalities. 
E.g., language 
is integrated with vision in 
Visual Question Answering \cite{antol2015vqa}, Visual Commonsense Reasoning \cite{zellers2019vcr}, Visual Dialogue \cite{visdial}, or Phrase Grounding \cite{plummer2015flickr30k}. Audio signal processing has made advances in speech recognition \cite{nassif2019speech} and (visual) speech synthesis \cite{alam2020survey}.
But ML applications may reach beyond modalities that are familiar to us: Astronomical and medical imaging techniques record wavelengths outside of what we call visible light. Genetic research measures signals alien to human perception, like activity and structure of molecules. Hence, we argue that current definitions of multimodality fall short of capturing the full space of multimodality in the ML era, and -- even more seriously -- that the field of multimodal ML, including vision and language integration, is lacking a proper definition of \emph{multimodality}.

\begin{figure}
    \centering
    \includegraphics[width=\linewidth]{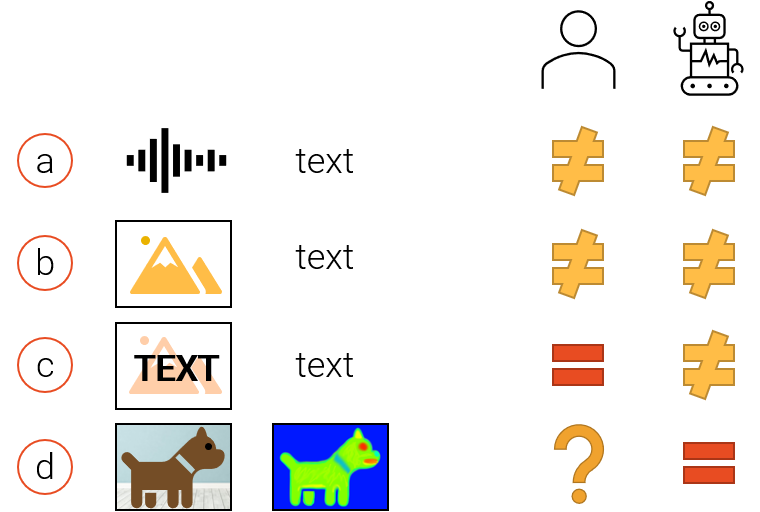}
    \caption{Are these examples instances of the same modality?  $\pmb{=}$ the same; $\pmb{\neq}$ different.  Depending on perspective, input data can be judged differently. Human- and machine-centered views would agree for (a) speech and text $\pmb{\neq}$, (b) images and text $\pmb{\neq}$. For (c), an image of text and text, the opinions could differ, while for (d), a visible light vs. infrared picture,  humans could not even judge the infrared data, since it is not within their sensory capability.
}
    \label{fig:one}
\end{figure}

The point we aim to make in this position paper can be crystallized by asking the following questions regarding the input types shown on the left of Figure\ \ref{fig:one}:
\emph{Are the data types shown on the left 
different from those that appear on the right, or are they instances of the same modality?} This question leads to different answers, depending on whether we take a human-centered as opposed to a machine-centered standpoint. 
In this position paper we reason that either standpoint is insufficient, therefore we develop a 
definition of multimodality that allows researchers to judge cases as displayed in Figure\ \ref{fig:one}, as involving or not different modalities from a \emph{task-specific perspective}.


Our contributions are three-fold:
\begin{enumerate}[label=\roman*), noitemsep] 
    \item We discuss existing definitions of multimodality and argue that they are insufficient to define multimodal tasks in the era of Machine Learning.
    \item We propose a task-relative definition of multimodality and illustrate it for a variety of (multimodal) tasks and data. 
    We argue that a task-relative definition better suits the current age of diversified data, where the field attempts to create multimodally enabled intelligent machines.
    \item By providing a novel definition of multimodality, we offer a foundation for building a road-map towards multimodality in NLP and beyond.
\end{enumerate}

\noindent
Multimodality is \emph{a new challenge in NLP} with research focusing on grounding language into other modalities. But how can we sensibly choose
other modalities to ground language in, if we 
are not clear about what kinds of modalities language can represent  itself,
and about what constitutes a modality in the first place? 
For building a roadmap towards Multimodality in NLP, we must establish common
ground for what multimodal ML is, what are possible instantiations or specializations of multimodality, and what research questions it gives rise to. 
Pinning down what makes a task multimodal in terms of information diversity, as we set out to do in this paper, is one important research question to ask. And from deeper understanding of what makes a task multimodal, we may -- conversely -- derive novel tasks.

\section{Multimodality and multimedia} \label{sec:multimedia}


It is difficult to write about multimodality in ML and not to trigger the question ``What about multimedia? Isn't multimodal machine learning actually just multimedia in machine learning?''  
But the use of the terms ``multimodality'' and ``multimedia'' in academic literature is very diverse. Attempts to compare and better capture these terms focus on different aspects and center on human society, human-to-human interaction and learning environments \cite{anastopoulou2001multimedia, bezemer2008writing, LAUER2009225}. We, however, focus on multimodal machine learning where the information receiver and processor is a ML system.

To decide whether \emph{multimodality} as used so far is the right 
term for the field of multimodal ML, we thus consult
general definitions for ``medium'' and ``modality'' 
in the Oxford Advanced Learner's Dictionary, rather than 
the academic literature. For ``medium'' we find:
\begin{enumerate}[label=(\alph*), noitemsep]
    \item \label{enum:media1} ``a way of communicating information, etc. to people''  comprising text, (moving) images, graphs, sounds.
    \item \label{enum:media2} ``a substance that something exists or grows in or that it travels through'' being mainly applied in natural sciences.
\end{enumerate}

\noindent
``Modality'' on the other hand is defined as follows:

\begin{enumerate}[label=(\Alph*), noitemsep]
    \item \label{enum:modality1} ``the particular way in which something exists, is experienced or is done''
    \item \label{enum:modality2} ``the kind of senses that the body uses to experience things'' being applied in biology.
\end{enumerate}

We argue that the term \emph{multimodality} should be preferred in the context of
ML, since multimodal ML aims to enhance 
world perception through
ML systems (cf.\
def.\ 
\ref{enum:modality1}). Finding new ways of presenting information to humans (cf. def. \ref{enum:media1}) is an endeavour that can benefit from multimodal ML while not being
the main focus of multimodal ML research. For the rest of the paper, we will use the term ``medium'' or ``media'' in different contexts, referring 
to the meaning defined in \ref{enum:media2}.

In the following Section \ref{sec:def_so_far}, we discuss how \emph{multimodality} has been interpreted by the multimodal ML field and highlight the shortcomings of the corresponding definitions in Section \ref{sec:against_existing}.

\section{How multimodality is defined -- so far} \label{sec:def_so_far}
In the multimodal ML literature and beyond, we find three general ways of defining ``modality'' or ``multimodality'': not at all or etymologically (bypassing the problem), or by way of 
a human-centered or a machine-centered definition\footnote{For the scope of this paper, we disregard the statistical sense of ``multimodality'', which describes a distribution with more than one peak. Such distributions can occur
with any kind of data, unimodal or multimodal in the sense of ``modality'' we use for this paper.}.

\paragraph{Not at all or etymological}
Especially recent publications, as in \citealp{lu202012,tan-bansal-2019-lxmert,gao2019multi}, bypass a definition, assuming that
the term 
is generally understood.
Others offer an etymological definition:
multimodal research involves not 
one, but \emph{multiple} modalities \cite{zhang2020multimodal}. Clearly, 
this definition leaves the notion of modality itself unexplicated.


\paragraph{Human-centered}
Popular definitions of multimodality rely on the human \textbf{perceptual} experience, as found in 
\citet{101109/TPAMI20182798607,lyons201618,ngiam2011multimodal,kress2010multimodality}. From this literature, we chose the following illustrative example, because 
the work focuses specifically on multimodality for ML, as is the interest of this paper:\footnote{Cf. section \ref{subsec:against_human} for considerations about other human-centered formulations.}
\begin{myquote}{1em}
\emph{
    ``Our experience of the world is multimodal -- we see objects, hear sounds, feel texture, smell odors, and taste flavors. Modality refers to the way in which something happens or is experienced''.} \hfill \citet{101109/TPAMI20182798607}
\end{myquote}

This view appeals to humans, who are bound to their senses when experiencing the world. It is thus an intuitive explanation of the concept of multimodality, focusing on the propagation channels that 
human communication is adapted to (e.g., vision, sound).

Using this definition, one can agree for Figure \ref{fig:one}.a that speech (hearing) and text (seeing) are different modalities. But decisions are less clear for
images and text as in Figure \ref{fig:one}.(b,c), as humans perceive both of them with their visual apparatus. Hence, as for written and depicted language, the human-centered definition \emph{contradicts} the common conception in the community, that vision and language are different modalities, as in \citet{lu2019vilbert, su2019vl}.

\paragraph{Machine-centered}
Another accepted perspective for defining multimodality is a machine-centered one, that focuses on the state in which information is transferred or encoded before being processed by a ML system:
\begin{myquote}{0em}
\emph{
    ``In the representation learning area, the word `modality' refers to a particular way or mechanism of encoding information.''} \hfill \cite{guo2019deep}
\end{myquote}

This definition is practical, focuses on the technical aspects of data representation, and captures how different
types of inputs usually require specific programming solutions.
For example, 
neural architectures typically use CNNs to encode
images (exploiting $2d$ patches) and LSTMs to encode text (modeling sequential reading), exploiting
the respective
architecture's inductive bias.
From this viewpoint, the machine-centered definition naturally regards
images vs.\ text
as different modalities (cf.\ Figure \ref{fig:one}). However, recent developments in neural architectures are challenging this view, since multimodal transformers are processing both images and text with transformer blocks \cite{lu202012, tan-bansal-2019-lxmert, su2019vl}.

\section{Why we need a better definition}\label{sec:against_existing}
In this section, we indicate shortcomings of the existing definitions and motivate why time is ripe for a new definition of multimodality.

\subsection{Human-centered} \label{subsec:against_human}
The human-centered definition 
(Section \ref{sec:def_so_far}) is rooted in research on how information is communicated 
to humans through, e.g.,\ GUIs
\cite{bernsen2008multimodality,jovanovic2015key,lyons201618}. However, this definition does not cover the large gap between the rich sensorial capabilities of humans as opposed 
to machines, 
and
leaves open
in which ways the signals that can be perceived by the variety of human senses 
 will be converted 
to specific types of inputs that can be given to the ML system.
Moreover, while we think that the human biology and psychology are and should be a valuable inspiration for ML research, we contend
that ML should look far beyond the human, into other organisms\footnote{e.g., a self-driving car imitating a nematode's nervous system, as in \citealp{lechner2020neural}}.

We will discuss three aspects of the gap between humans and machines: (a) the types of inputs their sensors can detect, (b) the range of these inputs, and (c) the importance of the processor -- a human or a machine -- 
that processes
the inputs, 
playing the role of
the  (multimodal) agent that determines how inputs are perceived.

\paragraph{(a) Input types}
Human senses are not prescriptive 
for
the sensorial apparatus of machines. Humans, on one side, are limited to the senses nature has gifted them with: sight, hearing, touch, etc.
Machines, on the other hand, are
only limited by the creativity of physicists or engineers: machines read tapes by measuring magnetization, or DVDs by detecting microscopic engravings.  What modalities would the human-centered definition assign to these signals? 
Also,  machines can 
surpass the human sensorial capabilities,
by performing more
exact measurements of e.g., humidity or pressure. This may not seem too relevant when restricting multimodality to the context of NLU, as language was developed to fit the human experience of the world. In other fields however, ML systems are used on signals that are completely outside of the perception of humans, e.g., predicting gene regulation based on chromatin accessibility and transcription in biology \cite{minnoye2020cross}.

\paragraph{(b) Input range}
Biological sensory detection systems are restricted to specific ranges of signals, and thus impose unnecessary limitations when applied to machines \cite{bernsen2008multimodality}. Machines are limited only by the borders of human engineering\footnote{engineering solutions which can be biologically inspired} and use manifold materials and physical phenomena that biological organisms adapted to their specific environment do not.
For example, humans can detect and interpret only a tiny part of the electromagnetic spectrum (380 to 700 nanometers, \citealp{starr2010biology}) and call it \emph{visible light}. But machines can detect and (if programmed) process the whole electromagnetic spectrum,
far beyond the visible light.
Humans cannot perceive ultra-violet light -- and hence 
this modality is non-existent to them because they cannot experience it.
This again mainly impacts fields apparently remote to NLP, e.g., medicine, where imaging techniques are employed that measure wavelengths far outside of the perceptive range of humans. But we desire that future systems can combine their experience of e.g., both the visible light and other wavelengths with natural language (to aid medical diagnosis, for example).  

\paragraph{(c) The processor: a human or an ML system?}
Humans have the innate ability of seeing objects, hearing sounds and delivering \emph{some} interpretation of these signals.
Machines, by contrast,
can be clever sensors and count photons on a semiconductor or measure air pressure oscillations -- yet without special programming they cannot interpret or derive information from the inputs.
Detected physical quantities are then mapped to
a voltage sequence interpreted as 0 if the voltage does not surpass a threshold, 1 if it does.

Since ultimately, behind all data encodings, 
there are just 0s and 1s waiting to be interpreted by a program, we argue that multimodal ML research should focus on these programs, and that a definition of multimodality should answer the question: What are the challenges that a program needs to address when it is 
exposed to a new modality, rather than more unimodal data?
In humans, evolution has already addressed this question, by specializing
sensory organs and brain areas to
the 
processing and interpretation 
of various
input types \cite{schlosser2018short}. Multimodal ML is not there yet.

\paragraph{Language -- humanity's stroke of genius}
It is generally accepted in the multimodal ML literature that vision is one modality and language the other, typically given in form of text
\cite{kafle2019challenges}. But humans \emph{hear} speech i.e., spoken language,
\emph{read} written language with their visual apparatus, \emph{see} signed languages, or \emph{feel} Braille. The upside of the human-centered definition is that it captures the plurality of media that 
support the transmission of language and 
accordingly, it can 
assign different modalities to 
such different manifestations of language. 

However, we are 
concerned with
multimodality in ML and there are important edge cases for which the human-centered definition is inadequate:
Does
a screenshot of a text editor that displays
the content of a .txt file containing a
sentence $s$ represent
a different modality than the data encoding
in the text file?
For a human, it is the same visual modality,
since in both cases $s$ is visually perceived. 
But a machine needs
very different programming in order to extract the same information about sentence $s$ from
an \emph{image} 
vs.\ the \emph{ASCII encoding} of $s$ stored in a .txt file.

Having raised this criticism of
the human-centered view on multimodality, it seems like the machine-centered view, focusing on the encoding of information, 
can offer
a more viable interpretation of multimodality.

\subsection{Machine-centered}
\paragraph{Data representation}
While data representations are a challenge \cite{bengio2013representation}, we argue that 
representations themselves should
not be the defining trait for multimodal learning.
For example, if we follow the machine-centered definition, an undirected unlabeled graph and its adjacency matrix 
have to be considered
different modalities because they are different ways of representing a network. Similarly,
PNG and JPEG are possible encodings of the same image and hence would be considered 
different modalities.
This interpretation seems
unintuitive for tasks like image recognition, where the image format does not play a big role and is usually homogenized in pre-processing. 

Still,
there are applications where the 4\textsuperscript{th} dimension of PNGs is useful for encoding transparency or depth maps, by using
the additional channel that PNG has compared to JPEG. Here we stand with a puzzle: Should data representation matter or not?

\paragraph{When does representation impact information?
}
Speech and handwriting are propagated through different media (e.g. air for speech, ink trail on paper for handwriting) and are 
represented differently in a computer -- time series of amplitudes for speech, images or textual transcriptions for handwriting. Does this 
make them \emph{different} modalities? Both speech and writing can be propagated through the same digital medium 
(fiber-optic cable) and if speech has been transcribed to text in pre-processing, they can 
have the same (ASCII) encoding. Does this
make them 
the \emph{same} modality? 

When converting between media and representations, information can get lost. But
without knowing the multimodal ML task and the information it requires, we 
cannot decide whether the loss was noteworthy or not. Below,
we will argue that crucial factors for defining multimodality are not only an efficient encoding of information, but also the ML task itself.

\section{A task-relative definition} \label{sec:multimodality_def}
We argue that a definition of multimodality for multimodal ML should relate to the task an ML system needs to solve,  
since it is the task that determines what \textbf{information} is relevant and in which 
\textbf{representation} this information can be efficiently stored.
The human- and machine-centered definitions try
to capture the 
essence of multimodality in a task-agnostic manner, relating it to categories of
human experience, media, representations, and encodings. As shown
in Section \ref{sec:against_existing}, these definitions turn out to be
insufficient in view of
the plurality of physical sensors, tasks and data types of the present. 

Instead, our task-relative definition aims to answer the question: 
Under which conditions does a multimodal setting reveal crucial insight \emph{for the given task}
-- and do we need  multiple  modalities?
In our view, (i) different inputs can contribute specific information, but (ii) what is relevant information can only be determined in relation to the task at hand; and only by taking the task into account (iii) we can determine the status of the inputs as (possibly complementary) modalities.

\subsection{Task-relative definition of multimodality} \label{subsec:our_def}
\begin{figure}
    \centering
    \includegraphics[width=0.9\linewidth]{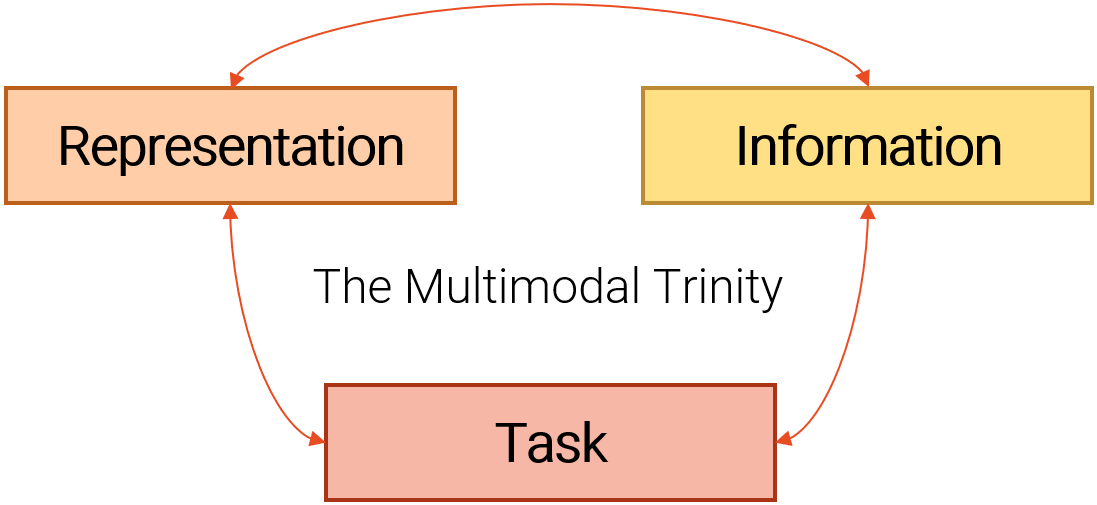}
    \caption{
    Our definition of \emph{multimodality} determines the  modalities of input channels  by considering i) how each input channel is  \emph{represented}, ii) whether the \emph{information} units each input carries is complementary to each other iii) \emph{in relation to the ML task.}}
    \label{fig:multimodal-trinity}
\end{figure}

We propose the following \textbf{task-relative definition} of multimodality in ML that relates \emph{representation}, \emph{information} and \emph{task} as depicted in Figure \ref{fig:multimodal-trinity}.

\begin{myquote}{1em}
    \emph{A machine learning \textbf{task} is multimodal when inputs or outputs are \textbf{represented} differently or are composed of distinct types of \textbf{atomic units of information}.}
\end{myquote}

\noindent
Note that our definition 
covers \emph{both} input and output to a system. In the following, we will primarily discuss and exemplify uni- and multimodal \emph{inputs}. The same arguments and examples apply symmetrically for \emph{outputs}.

Since the definition focuses on multimodal ML, when considering inputs, the
representation of interest is the direct input to the learning system; any prior representational steps that are not seen, or filtered by pre-processing, are irrelevant for our definition. The reverse applies to outputs and post-processing.

\begin{figure*}
    \centering
    \includegraphics[width=\linewidth]{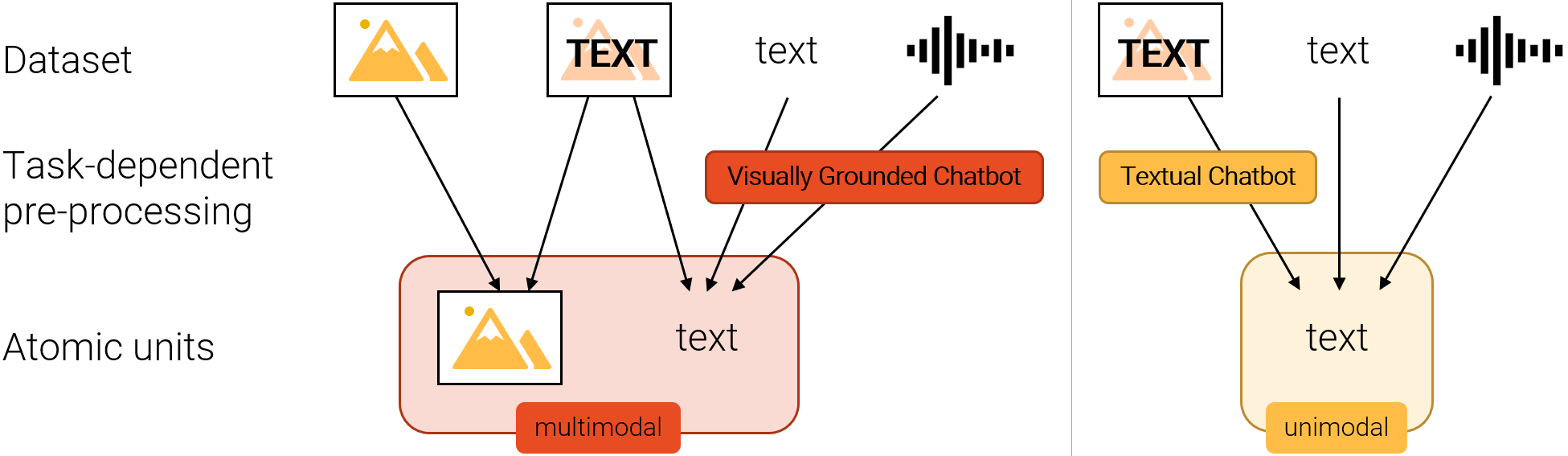}
    \caption{The task defines what are the required atomic units of information for solving it. A task that requires multiple atomic units after pre-processing is multimodal.}
    \label{fig:representations}
\end{figure*}

\paragraph{Atomic units of information}

\emph{Information} is one of the three dimensions we use to decide multimodality. We work with the information-theoretic interpretation of information that measures
the decrease of uncertainty for a receiver \cite{shannon1948mathematical}. In our case, the receiver is the ML system -- whose objective is to solve a certain task.
Hence we ask: What information does the input carry that is relevant for (or decreases uncertainty of) the task solution? As a defining criterion on whether two input channels contribute information from different modalities, we consider what types of atomic units of information they provide: they are of different types, if they cannot be captured by a 1-to-1 mapping between their domains. In this case, we speak of multiple modalities.

In other words, if
we find that \textbf{i)} after pre-processing, inputs $x_1$ and $x_2$
are represented the same,
we examine whether they live in the same modality or not by asking whether $x_1$ and $x_2$ are formed of different atomic units: If \textbf{ii)} we can't establish a
bijective mapping between the domains of $x_1$ and $x_2$, then $x_1$ and $x_2$ must be composed of different atomic units, and even if \textbf{i)} 
was found to be the case, they represent different modalities.

Similarly, if \textbf{iii)} after pre-processing, $x_1$ and $x_2$ are represented \emph{differently}, the task is multimodal.

The bijectivity criterion is lenient towards a task-dependent error for real-world applicability. Regardless of multimodal characteristics, low data quality and compression can cause loss of task-relevant information.

\paragraph{Information and its atomic units}

Atomic units of information are not to be confused with information itself. In analogy, atomic units of information are to information what
meters (unit of measuring space) are to objects of a
certain size. Types of atomic units differ between each other like the meter (space) differs from the second (time). More concretely, atomic units of information apply to the \emph{data domain}; information itself applies at the \emph{data sample level} and can be accumulated by sampling more data (e.g. adding images of cats to a set of dog images). While it is impossible to have an 1-to-1 mapping between cats and dogs (information adds something previously unknown), our definition proposes bijective mappings between information \emph{units} to identify modality-specific but data sample-invariant properties, e.g. edges and texture for images, time directionality for video, (abstract) concepts in text, nodes in graphs, etc.

We thus speak of a new modality when it contributes information that cannot be delivered by larger (but not infinite) amounts of unimodal data.
Note that the same information captured in one modality may be encoded in a different modality, however, not necessarily with the same efficiency:
We can, in infinite time, describe every minute detail of a landscape unimodally through language. But it is clearly more efficient to capture the details of a landscape in a different modality, e.g., 
a photograph. In general, any kind of information can be reduced to a string of 1s and 0s, yet, depending on the information source and the given task, another representation might be more convenient.

\paragraph{Task-specificity}
The \emph{task} is the second crucial characteristic of multimodality because 
it determines (i) what information 
from
the input 
is necessary to solve the task as discussed above, and (ii) it determines two crucial components: \emph{input and output} and how they relate. Regarding inputs and their possible (multi)modality, we critically focus on the nature of the data that forms the \emph{direct input} to the ML system,  \emph{disregarding} any data representations that the input may take during any pre-processing stages.
Hence, we draw a sharp line between input encodings and representations that are part of the learning process, and that will be continuously refined -- as opposed to input formats that are external to the learning process.

For our task-dependent definition of multimodality it will therefore not matter whether the original input was speech, but was transcribed to text \emph{if}, for example, the task at hand does not require information that is specific to spoken language and can be conveyed by a textual representation of spoken input (see Figure \ref{fig:representations}). Also for image recognition it is not crucial whether images come as PNG or JPEG, and harmonizing the data is a matter of pre-processing.

\paragraph{Atomic units vs. data representation}
The term \emph{data representation}, as we use it here, refers to the encoding of information and data formats. However, \emph{atomic units} are not bound by their technical implementation -- they constitute the informational content in \emph{relation} to the task. For example, PNG and JPEG are undeniably different data representations which can, depending on the task, represent 
(a) two \emph{different} types of atomic units of information, if the additional dimension in PNG is important for the task (e.g., in view of encoding transparency, or depth), or (b) a \emph{single}
type of atomic units, because a task-specific bijective mapping can be established 
that does not lose (task-relevant) information.

\paragraph{Not sensor-specific} 
Our definition is especially robust to changes of medium and representation of information via data transfer or storage. We can thus neglect the physical or biological sensors that capture the data, and the encoding, transmission or storage of the data until it reaches the processor. This property makes our task-relative definition a robust definition: the constant change of representation that information may undergo, does not immediately span a new modality.

\subsection{Applying the task-relative definition} \label{subsec:def_applied}

We now apply our definition of multimodality to various examples -- including edge cases -- in order to demonstrate its breath, flexibility and robustness.

\subsubsection{Images vs. text}
By our definition, tasks working with images (stacks of intensity matrices) and text in ASCII format are multimodal, because they consist of different atomic units of information:
An image can be truthfully described by multiple textual descriptions, and similarly, text translates equivocally to images (e.g., pictures of different hand writings).
The decision becomes less clear-cut, if we are given
two inputs:
natural images and images of text. Both inputs are intensity matrices and therefore unimodal. However, if the task does not consider the differences in
hand writing style 
and applies Optical Character Recognition on the images of text to obtain e.g. an ASCII text representation,
the 
\emph{images of text} turn into \emph{text} in pre-processing. These, alongside the images, make the setting multimodal. In the next paragraph we will show examples of both of these approaches -- using images of text unimodally, or using OCR to convert the images of text to the two modalities of images \emph{and} text.

\begin{figure}
    \centering
    \includegraphics[width=\linewidth]{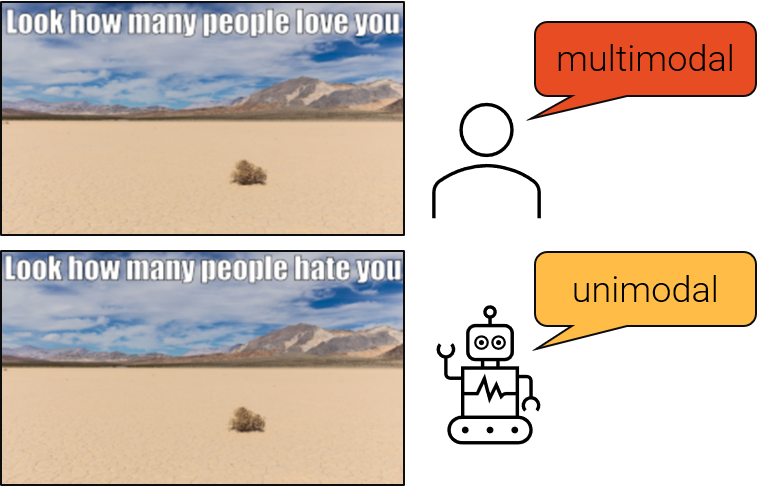}
    \caption{Two memes from the Hateful Meme Challenge \cite{kiela2020hateful}. The rendered text can change the meaning of the picture and -- from a human perspective -- constitutes the multimodal aspect of the problem. But an ML system, tasked to extract meaning from both the rendered text and the depiction of the desert, has to start from raw pixels in both cases.}
    \label{fig:meme-challenge}
\end{figure}

\paragraph{Examples concerning images of text}
For us as humans, it is tempting to see an image with text and think of it immediately as multimodal. But we experience text through our visual apparatus. For ML systems, images are always stacks of pixel matrices. Some
images contain pixels that can be read by humans as text, others only contain pixels that depict an object -- but the unit of information stays the same. To disentangle the issue of how images, text, and images of text relate to multimodality, we give two examples of multimodal models -- winning models of the Hateful Meme Challenge dataset and CLIP.

The Hateful Meme Challenge dataset \cite{kiela2020hateful} consists of memes -- natural images with \emph{text rendered into the image} as depicted in Figure \ref{fig:meme-challenge}. The task is to classify whether the memes are hateful or not.
For now, winning systems \cite{zhu2020enhance, zhong2020classification} are using both the meme image and a string representation of the meme's text besides the image. Thus, they solve the task multimodally, because text in string format can not be mapped 1-to-1 to the meme text, which could vary in font, color, or size.
But \emph{ideally}, models would be able to extract all relevant information only from the meme image -- like humans do -- without the help of an additional string representation of the meme text, therefore solving the task unimodally.\af{\footnote{At first sight: internally, humans know how to map text displayed in an image to a textual representation, thus turning images to text on the fly. ML systems are not yet there, but they are getting closer, as the example of CLIP will show in the next paragraph.}}

\begin{figure}
    \centering
    \includegraphics[width=.85\linewidth]{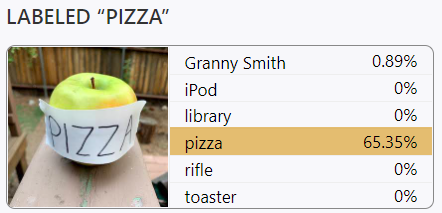}
    \caption{In this example, \citet{goh2021multimodal} show that CLIP can detect a high similarity between the image containing a handwritten note reading ``PIZZA'' and the textual input ``pizza'', i.e. CLIP has learned to \emph{read from images} like humans can do.}
    \label{fig:clip-attack}
\end{figure}

CLIP \cite{radford2021learning} is a multimodal model with impressive zero-shot applications in image recognition. It is trained to predict similarity scores between image-text pairs. Some of the images in the training data also show characters, words or sentences, and because CLIP was trained with these images and their corresponding OCR data, CLIP has learned to correctly identify what the \emph{pixels depicting text} represent (see Figure \ref{fig:clip-attack}). CLIP is multimodal because its inputs are an image and text, for which it has to predict the similarity; whether or not the image shows text is irrelevant.

\subsubsection{Images of infrared vs.\ visible light}
Infrared images and visible light images are represented uniformly (a stacked grid of photon counts of different energy intervals). Additionally, we can define a bijective mapping between the two, e.g. by adding/subtracting a fixed frequency and thereby shifting the infrared image into a visible light image (as used for night vision). Therefore, by our definition, they represent the same modality. In this way, our definition enables us to define modalities for information not directly perceptible by humans because of sensory limitations. For example, the photons of infrared light do not carry the correct amount of energy to be perceptible by our eyes, as it is the case for
most of the electromagnetic spectrum.

\subsubsection{Language}
Finally, our definition captures two key traits of the multimodal nature of language: (i) coming in many forms (speech, handwriting, signed language, ASCII-code), language constantly switches representations and media. In cases where (ii) after pre-processing different language representations 
cannot be converted one to the other without losing task-relevant information (e.g., intonation, hesitation, modulation in speech), they become multiple modalities, like speech--text, or handwriting--text, etc.

With our definition, languages like English and Japanese are considered to be unimodal if  after pre-processing both are represented in Unicode. If not, handling them both
becomes a multimodal task. This behavior relates the essence of multimodal ML and multilingual NLP, in terms of their complementarity: 
There are concepts in some languages that cannot be efficiently translated to other languages; much like humans cannot conceive how a bee sees ultra-violet light \cite{chittka2004color}.  

\subsection{Relevant aspects of multimodality} \label{subsec:rel_aspects}
Similarly to previous definitions of multimodality, we focus on the representation and content of inputs (or output, for that matter), but we consider
them in \emph{relation} to the \emph{task} (see Figure \ref{fig:multimodal-trinity}):

\begin{enumerate}[label=(\roman*), noitemsep]
    \item Starting from the \emph{task}, we capture \emph{what information is put to use} and relevant for solving it. Hence, despite the presence of e.g. video and language in a dataset, the task may be such that one of them is not relevant and could be fully ignored.
    \item We focus on the \emph{representation} of the information that forms the direct input to the ML system after pre-processing.
    Unlike previous definitions, we reduce the importance of the medium in multimodal ML. The medium may change without modifying the nature of the \emph{information} that is represented, used and needed for the task.
    \item Finally and crucially, we determine the uni- vs. multi-modality of the information joined to solve a task, by analyzing its lack of bijectivity, i.e. its \emph{complementarity}.
\end{enumerate}

Hereby we conclude that it is only in the context of a task and data used to solve it that we can analyze the issue of information relevance and bijectivity/complementarity, i.e. of \emph{multi}modality. In this sense we argue that there is no multimodal input, output, or data \emph{per se}; it is only through a task that requires multimodal information sources to be solved that the corresponding inputs or outputs can be truly considered multimodal.

\section{Conclusion}
In this paper we have shown how human-centered definitions of multimodality are too restrictive for current and future developments of ML. Also, machine-centered definitions focusing on representations only do not capture the crucial trait of multimodal machine learning. We 
instead propose
a new definition of multimodality that focuses
on the \emph{relation} between representations, information and the given task, and that -- through the novel dimension of the \emph{task} -- enables us to make much sharper distinctions compared to current standards, while covering a much wider 
spectrum of multimodal 
data.

With this position paper, we (a) offer
a working definition on how to use the term \emph{multimodality}, (b) aim to raise awareness that defining multimodality is harder than expected, and (c) invite the community to \emph{discuss} these challenges and (why not?) to provide a better definition.

\section{Ethical considerations}
In the present paper, we portray the opinion that the human experience of the world should not be normative for defining the multimodal characteristic of a machine learning system. Instead, we
claim that machine learning should draw inspiration from human biology and psychology, but not limit itself by imitation. Independently of (a) where the design of machine learning systems is inspired from, and (b) their 
capabilities, which may extend or exceed those of humans, we strongly believe that machine learning should be done for humans' service, following the ethical considerations developed and accepted by human society.

\section*{Acknowledgements}
We thank our anonymous reviewers for their insightful questions and comments. This theoretical research and question about how to define multimodality was inspired by fruitful discussions in the Copenhagen meeting of the Multi3Generation COST Action CA18231.

\bibliography{anthology,my}
\bibliographystyle{acl_natbib}




\end{document}